\documentclass{svproc}
\usepackage{graphicx}%
\usepackage{multirow}%
\usepackage{amsmath,amssymb,amsfonts}%
\usepackage{mathrsfs}%
\usepackage[title]{appendix}%
\usepackage{xcolor}%
\usepackage{textcomp}%
\usepackage{manyfoot}%
\usepackage{booktabs}%
\usepackage{algorithm}%
\usepackage{algorithmicx}%
\usepackage{algpseudocode}%
\usepackage{listings}%
\usepackage{makecell}
\usepackage{subcaption}
\usepackage{url}
\usepackage{mdframed}

\begin{document}
\mainmatter              

\title{
\includegraphics[width=1\textwidth]{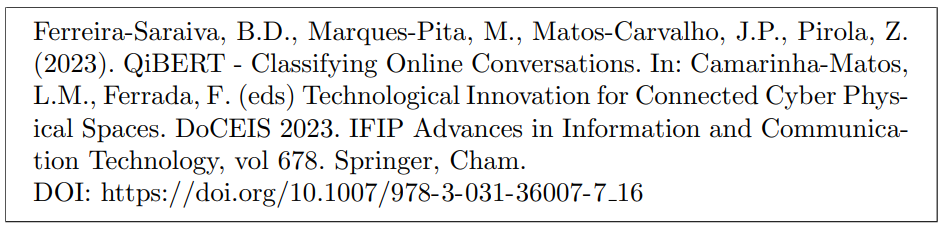}~ 
\\[1cm]
QiBERT - Classifying Online Conversations Messages with BERT as a Feature}
\titlerunning{QiBERT}  
%
\author{Bruno D. Ferreira-Saraiva\inst{1,2} \and Zuil Pirola\inst{2} \and
João P. Matos-Carvalho\inst{1,2} \and Manuel Marques-Pita\inst{1,2}}
\authorrunning{B. D. Ferreira-Saraiva et al.}
%
\institute{COPELABS, Universidade Lusófona, Campo Grande 376, 1749 - 024 Lisboa, Portugal\\ 
\and
CICANT, Universidade Lusófona, Campo Grande 376, 1749-024 Lisboa, Portugal\\
\email{\{bruno.saraiva, zuil.pirola, joao.matos.carvalho, manuel.pita\}@ulusofona.pt}}

\maketitle              

\begin{abstract}
Recent developments in online communication and their usage in everyday life have caused an explosion in the amount of a new genre of text data, short text.
Thus, the need to classify this type of text based on its content has a significant implication in many areas.
Online debates are no exception, once these provide access to information about opinions, positions and preferences of its users.
This paper aims to use data obtained from online social conversations in Portuguese schools (short text) to observe behavioural trends and to see if students remain engaged in the discussion when stimulated.
This project used the state of the art (SoA) Machine Learning (ML) algorithms and methods, through BERT based models to classify if utterances are in or out of the debate subject.
%
%
Using SBERT embeddings as a feature, with supervised learning, the proposed model achieved results above 0.95 average accuracy for classifying online messages.
Such improvements can help social scientists better understand human communication, behaviour, discussion and persuasion.

\end{abstract}
\section{Introduction}
Influenced by social networks based on short and fast content such as Twitter and Tiktok, the digitalized post-pandemic school can adapt, emulating these types of networks and motivating the participation of students in the discussion of current topics \cite{Careaga}.
One of the possibilities is the use of multi-participant chat, a form of chat with several participants talking synchronously through textual communication \cite{Uthus2013}.
Chats, and their integration with teaching, have already been studied \cite{Anjewierden2007,Trausan} Despite their implementation advantages, these are increasingly being incorporated into the range of teaching tools, and it is important to know whether or not students are engaged with the themes proposed.

Natural Language Processing (NLP) tools can assist in the analysis and even in the classification of these data as useful or not useful. Recently, several studies have investigated the classification of short texts \cite{Alsmadi2019,Danilov2021,Demirsoz2017,Hu2022,Lee2016}.

However, conventional text classification tools are not directly suited to this type of medium with short texts \cite{Uthus2013}. This inadequacy is mainly due to the characteristic difference between the two types of text. Short texts mainly present sparsity, ambiguity, shortness and incompleteness.

In general, those studies follow the conventional classification pipeline containing four levels: Features extraction, Dimensionality Reduction, Classification Techniques and Evaluation \cite{Kowsari2019}.

For this study, we intend to replicate this pipeline and focus on the analysis of chat conversations in order to understand whether or not the students are talking about the subject they were stimulated. 

The goal is to be able to classify the messages as “on the subject” or “off the subject”. For that, We will use multilingual BERT models \cite{Devlin2018,Lin2020} trained in multiple languages, including Portuguese (European), which will be applied to sentence units via SBERT \cite{sbert}. It is expected that these models can effectively capture the semantics of the analysed messages with the least amount of training data possible.

This paper is organised as follows: Section II describes the state-of-the-art (SoA). Sections III and IV will present the characteristics and the way in which the analysed data were acquired, including the annotation procedure. In section V the proposed method will be presented, in section VI the results are shown discussed and in section VII the conclusions and the future work are presented.

\section{Contribution to Connected Cyber Physical Spaces}
Online debates are crucial to providing important data for the interpretation and classification of ML models. In this study, it was analyzed whether a given debate subject was maintained throughout the conversation, and to this end, different feature extraction models and ML algorithms were studied to classify whether or not the subject in question was discussed during the conversation. The presented work focuses on Connected Cyber Physical Spaces, especially on intelligent NLP models developed through ML algorithms.

\section{Related Work}

Online conversations come in many different formats.
There are studies on data such as discussion forums \cite{Hidey2017}, specific messages from platforms such as Facebook \cite{Meredith2014,xuan2019}, or Twitter \cite{Demirsoz2017}.
Although all these studies involve short texts, they have significantly different structures.
Replies to a tweet may come shortly after it is posted, but they can be made days later.
Discussion forums can last for years and have features like quotes and replies.
Among these and other differences, this paper focuses specifically on chats, where all participants in the conversation are simultaneously exposed to a virtual environment to discuss, in our case, the topic of racism.
The analysis of multi-participant chats, their problems, and their relationship to computational techniques has been widely studied \cite{Uthus2013}.
Computational models can even help social studies through Conversation Analysis (CA) \cite{Jucker2021,Meredith2019,Paulus2016}.
However, for this, it is crucial to understand what the participants are talking about in order to improve the reading and perception of the messages sent.
We intend to explore the classification of chat messages.
Short texts in chat rooms could have a few words, presence of abbreviations, spelling errors, or texts being supplemented in subsequent messages. 
All these characteristics, in addition to other factors, make feature extraction difficult. As a solution, the authors complement the short text with external knowledge.
Liu \cite{Liu2022} used external knowledge to enrich the semantic representations to develop a model based on TCN and CNN.
Hu \cite{Hu2022}, augmented the vector representations of the text by combining information from the message actors to generate mental features.
%

%
Danilov~\cite{Danilov2021} proposed 27 parameterised PubMedBERT options and new models for classifying academic texts.
There was also the use of BERT to classify political discourse~\cite{Gupta} in short texts on Facebook and Twitter.
With the application of BERT and other vector representations (Glove), Khatri~\cite{Khatri2020} used binary classification to classify sarcasm in tweets.
BERT was also used to create a graph convolutional network for classifying short texts~\cite{Ye2020}.

Motivated by the discussions above, in this work we aim to classify text messages present in a chat using conventional classification techniques and contribute to the discussion as follows:
\begin{itemize}
    \item It is possible to classify texts from chat messages, even if they only have short features (short texts);
    \item With a small amount of training data, supervised learning models have high accuracy;
    \item Using pre-trained BERT models in combination with the sentence embedding framework (SBERT) to train a robust sentence classification model.
    \item Use of feature selections to reduce the dimensionality of the model inputs
\end{itemize}

\section{Data Gathering}
The data for this research was collected from instant multi-participant messaging chat under the project "Debaqi - Factors for promoting dialogue and healthy behaviours in online school communities". Users were placed to debate in a private virtual environment and interactions were synchronous where any participant can contribute to the conversation at any time.

\begin{table}
\begin{center}
\def\arraystretch{1.5}
\begin{tabular}{l||c}
\toprule
\multicolumn{1}{c||}{\textbf{Item}} & \textbf{Number} \\ \hline
Rooms                               & 25.0          \\ \hline
Messages with Moderator             & 5303.0        \\ \hline
Messages without Moderator          & 4044.0        \\ \hline
Users                               & 309.0         \\ \hline \hline
Average Messages Length (chars)     & 61.4          \\ \hline
Median Messages Length (chars)      & 28.0          \\ \hline
Average Messages Length (tokens)    & 10.1          \\ \hline
Median Messages Length (tokens)     & 5.0           \\ \bottomrule
\end{tabular}
\end{center}
\caption{\label{table1}Data summary.}
\end{table}

The online conversations took place in a virtual environment involving Portuguese state high schools. There were 25 rooms, with 309 participants. The messages sent are predominantly short-text and have a median of five tokens (Table~\ref{table1}). The participant's ages were between 15 and 19 and we obtained previous consent from their parents for them to participate in the chat room debates. The students may or may not know each other and the chat application guarantees the anonymity of participants. Platform anonymity means that participants know that something was said by a particular user, but they do not know who the user is in the school context.

At the beginning of the conversation, students were stimulated through a video and the moderator also contributed through questions launched at a given time according to a moderation script. There is no way to set a certain conversation path or set a certain topic, so there is the possibility of students following the theme, changing the theme, creating sub-themes or even ignoring the proposed theme in order to boycott.

\section{Annotation}
In supervised models, as foreseen in this work, the classification model demands annotated data. 
The most convenient way to generate this annotated data is to use annotators that do it manually. 
It is important to define an annotation method that guarantees good inter-agreement \cite{landis,Krippendorff2016} between annotators and that can reliably transmit the annotated data to the classifier.
The annotation criteria used in this work were:
\begin{itemize}
\item Label 1 - Messages that were about the topic/subject \emph{``Racismo e Esteriótipos''}. Sentences containing words such as ”racism”, ”racist”, ”stereo-types”, ”culture”, ”prejudice”, ”black/white” were considered, as well as sentences like ”we are all equal/human”;
\item Label 0 - All messages that do not have a defined subject like greetings ("good morning", "hi"...), agreements/disagreements ("yes", "no", "agree", "disagree", "maybe"). All messages that have a defined subject, but are not directly linked to {``Racismo e Esteriótipos''} topic.
\end{itemize}

In a pilot annotation phase, only two rooms were randomly chosen and assigned to 3 annotators, where we obtained an average inter-agreement above 0.7 of Krippendorff's alpha as expected \cite{landis}. 
Therefore, the simple annotation criteria proposed was well understood among the annotators and can be implemented in the total pool of rooms. Despite having access to the entire conversational sequence of messages, annotators do not consider the context.
There may be messages that talk about racism or that were related (reply or quote) to a message about racism but do not necessarily have words that cite the topic directly. 
In this case, they were not annotated as messages of racism.
In the second annotation phase, the other 23 rooms were submitted to the same three annotators. The average \emph{Krippendorff's alpha} value for the three annotators, at the 25 rooms, was 0.77.

\section{Proposed Method}

In this section, we describe the methods used to build the models for the online text classification task.

\subsection{Feature Extraction}
\label{subsec:feature extraction}

\begin{figure}[h!]
	\centering
	\includegraphics[width=0.8\linewidth]{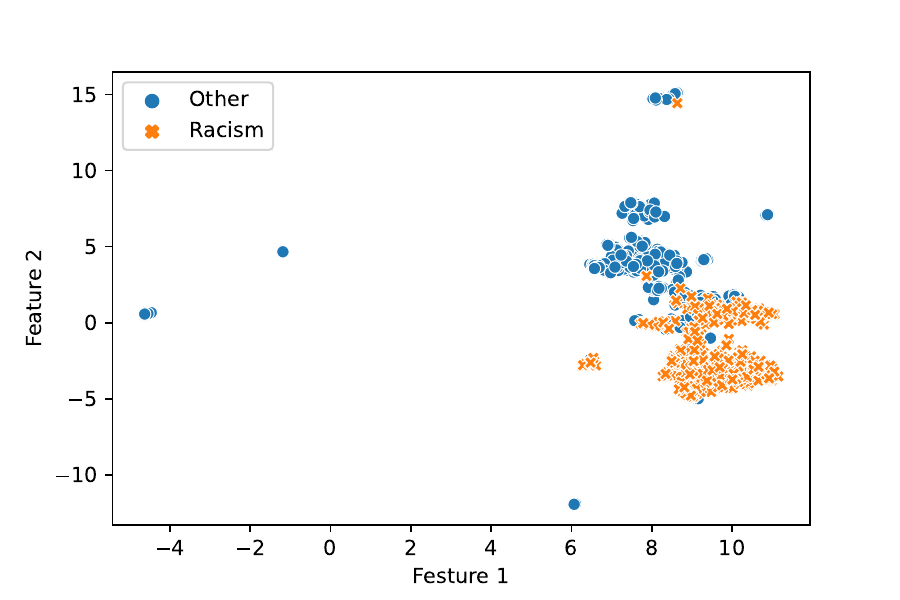}
    \caption{Comparison between ``Racism'' and ``Other'' subject sentence embeddings.}
    \smallskip
    \small
    {To be possible to visualise the sentences embeddings, UMAP~\cite{umap} was used to reduce the vector's 768 dimensions for two.}
	\label{fig:embeddings} 
\end{figure}

In the model-building process, feature extraction is crucial.
Word and sentence embeddings are commonly used to represent language features in the field of Natural Language Processing (NLP)\cite{mikolov,Bert,levy}.
Sentence embedding refers to a group of feature learning techniques used in NLP to map words or sentences from a lexicon to vectors of real numbers.
For the feature extraction stage of our study, we used embeddings from a pre-trained model~\footnote{SBERT Model: \textit{paraphrase-multilingual-mpnet-base-v2}. 	Multi-lingual model of paraphrase-mpnet-base-v2, extended to 50+ languages.
\url{https://huggingface.co/sentence-transformers/paraphrase-mpnet-base-v2}}
from SBERT framework \cite{sbert}.
SBERT outperformed the previous (SoA) models for all common semantic textual similarity tasks since it produces sentence embeddings, so there is no need to perform a whole inference computation for every sentence-pair comparison.
This framework is a useful tool for generating sentence embeddings where each sentence is represented as a size 768 vector (Figure ~\ref{fig:embeddings}).
This embeddings are based on BERT~\cite{Bert}, so they are contextual.
Once the embeddings were extracted for the training data, the sequence of embeddings was ready to feed the machine learning models.

\subsection{Training and Predictions}

After extracting the sentences embeddings we followed into two different approaches: 
\begin{itemize}
    \item We trained the six ML algorithms, mentioned above, with the raw embeddings (768 features);
    \item A feature selection method~\cite{stoppiglia2003ranking} was used in order to remove the less important features, before training the algorithms.
\end{itemize}

\begin{figure}[!h]
    \centering
    \includegraphics[width=0.8\linewidth]{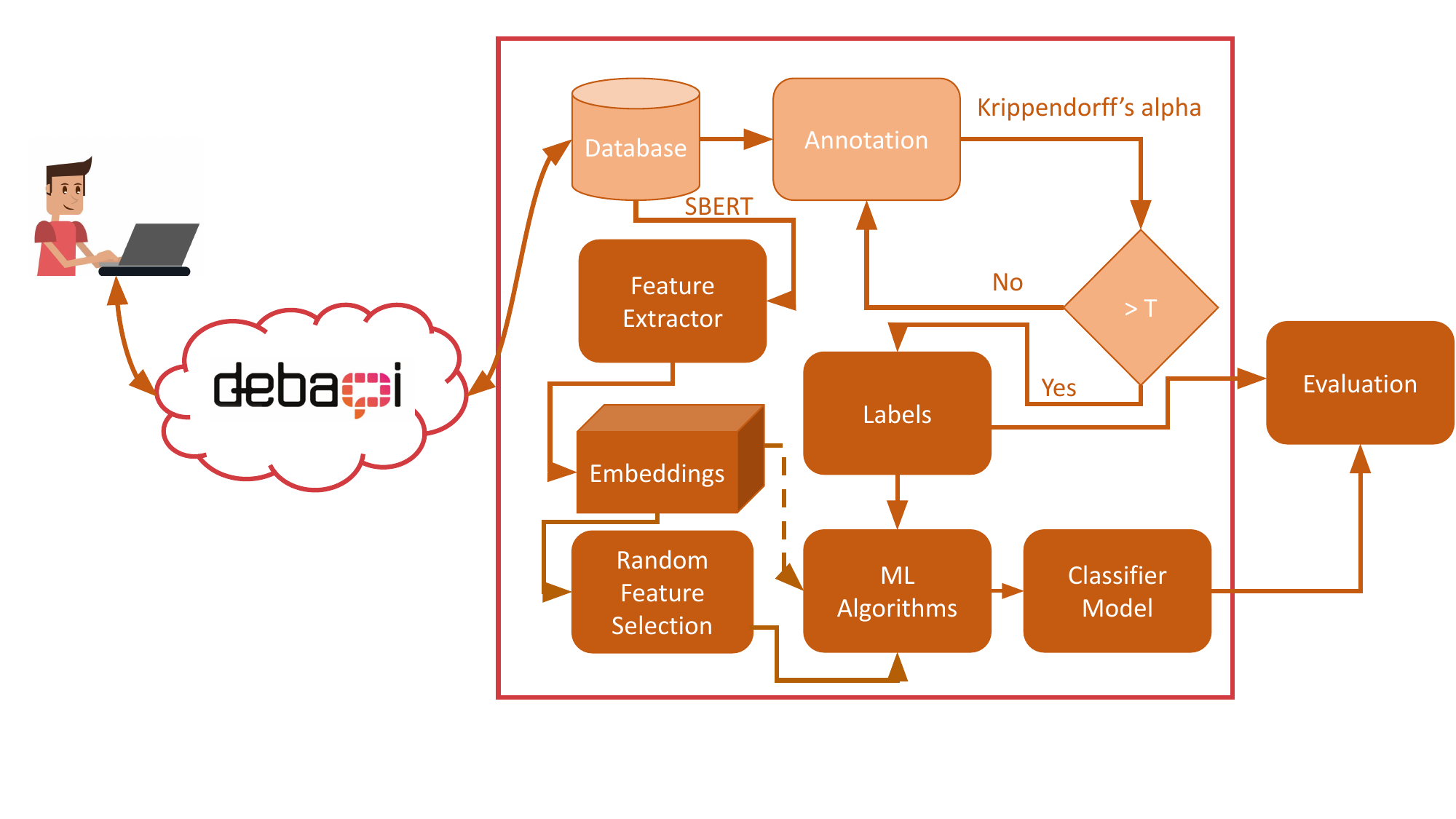}
    \caption{Proposed system architecture.}
	\label{fig:pipeline}
\end{figure}

For the first approach, we follow the dashed path of the pipeline (Figure~\ref{fig:pipeline}), where after extracting the embeddings, we move directly to the classification algorithms.

\begin{figure}[h!]
    \centering
    \includegraphics[width=0.7\linewidth]{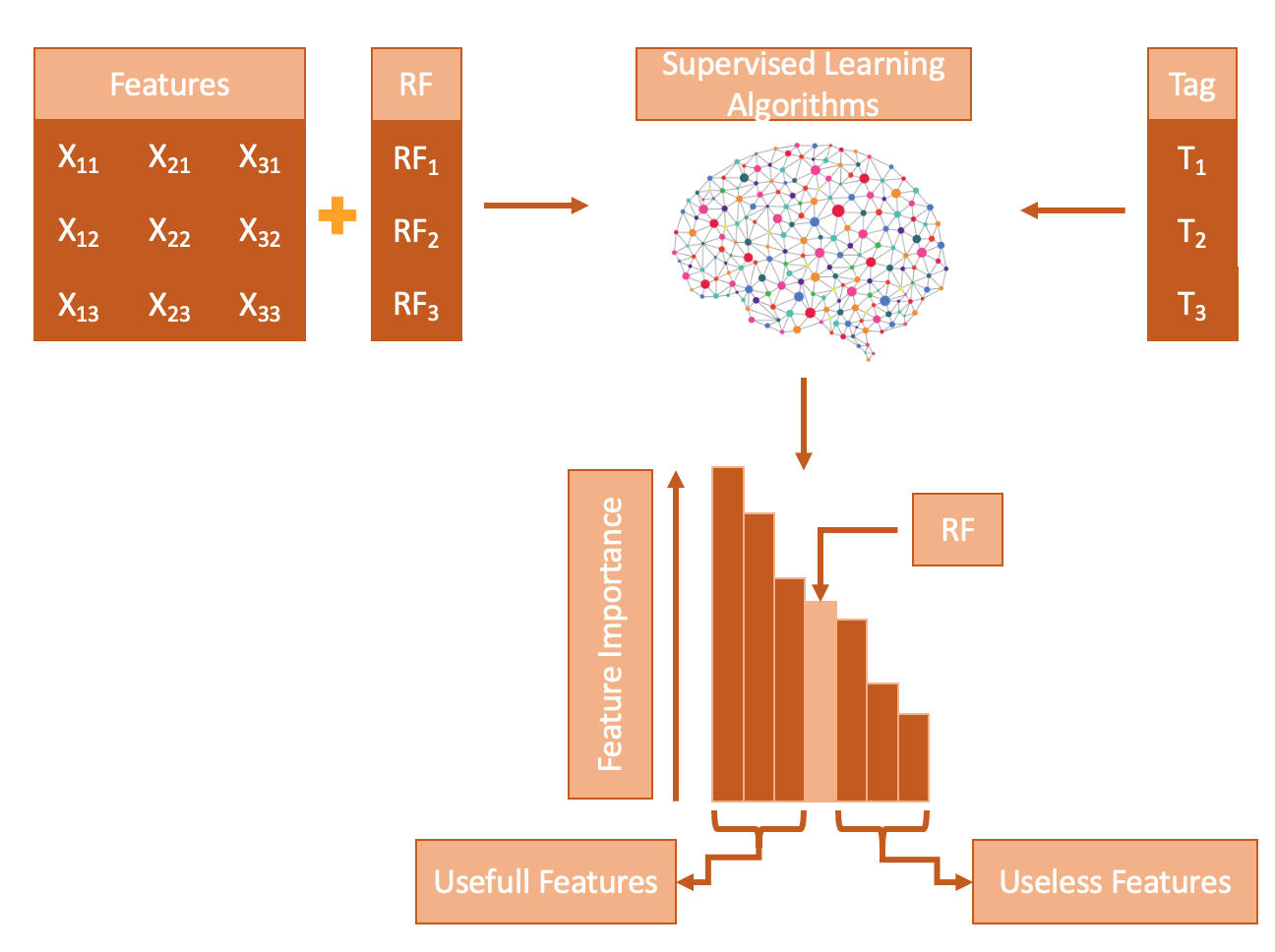}
    \caption{Feature selector method. Adapted from~\cite{stoppiglia2003ranking}.}
	\label{fig:random_feature}
\end{figure}

For the second approach, we have an intermediate step, before moving towards the classification algorithms stage, that aims to select the most important features (blue path of pipeline), in order to try to reduce memory usage as much as possible and maintain the proposed system performance.
For that purpose, given the sentences embeddings, created before, known as $X$ and a target (annotation) $Y$, a random vector $V$ was created and appended as a new feature of X: $X' = [X,V]$.
Now, with that data (X', Y), the next step was to train a Supervised Learning algorithm with with a relevant feature importance measure (Figure ~\ref{fig:random_feature}).
Generally, importance provides a score that indicates how useful or valuable each feature was in the model's construction and is calculated explicitly for each attribute in the provided dataset, allowing attributes to be ranked and compared to each other. 
Thus, XGBoost~\cite{xgboost_Chen_2016} was used in order to calculate the importance. 
If a given feature has a lower feature importance than the random feature, we set it as useless feature and therefore removed it from the original sentences embeddings~\cite{stoppiglia2003ranking}.
%
We ran 1000 Monte Carlo (MC) simulations and achieved, on average, around 85\% of feature reduction from the original dataset features (embeddings).

The next step was to train different machine learning models to create a classifier that can predict the class of test samples.
Classifiers such as Logistic Regression (LR), Support Vector Machine (SVM), Gaussian Naive Bayes (GNB), Bernoulli Naive Bayes (BNB), K-Nearest Neighbours (KNN), XGBoost (XGB), and Multi-layer Perceptron (MLP) were used \cite{9849888,svm,2019}.
Scikit-learn library was used to train these models.
Sentence embeddings were obtained for the test dataset in the same way as mentioned before. This way, they got ready for predictions.

\section{Experiments and Results}
\label{sec:Results}

The first developed experiment was directly related to the data annotation. 
Since there were three annotators, it was expected that they would not always agree, and for that reason two different models were investigated:
\begin{itemize}
    \item A model in which all three annotators agreed (Complete Agreement (CAg)).
    \item A model in which at least two of the annotators agreed (Majority Agreement (MAg)).
\end{itemize}

For the CAg model there was a total of 2334 messages annotated in concordance, from which only 17\% were annotated with a value ``1'' that indicated that the students sentences were explicitly addressing the subject of ``\textit{Racismo e Esteriótipos}'', as explained before.
For the MAg model, there was a total of 3727 annotated messages in which at least two of the annotators were in agreement. 18\% out of those had been annotated with the value ``1''.
Thus, in order for the data not to be biased, only 790 sentences were used for CAg model training and 1300 for MAg model training. This way, around 50\% of the messages would have been annotated with value ``1'' and the remaining 50\% were annotated with zero.

For this experiment, the ML algorithm choice was not the true focus. The main goal was to realise that annotated data, in which three of the annotators were in agreement, obtained better results than with only two annotators in agreement.
SVM was the selected ML algorithm and the training and testing sizes were set to 66\% and 33\%, respectively.
\begin{figure}[h!]
    \centering
    \includegraphics[width=0.6\linewidth]{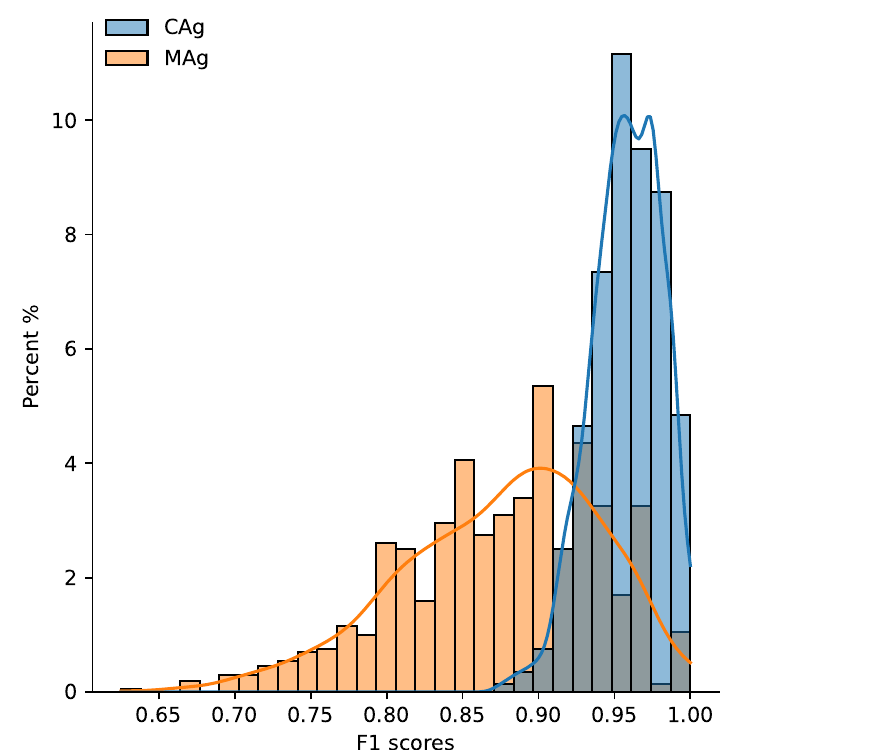}
    \caption{Comparison between Complete (CAg) and Majority (MAg) Agreement.}
	\label{fig:comp_100_maj_1000MC}
\end{figure}

As it can be seen in Figure~\ref{fig:comp_100_maj_1000MC}, we can intuitively perceive that the model CAg was much superior in the evaluation metrics chosen by the authors, than the MAg model.
The former obtained an average of 0.96 F1-score, whereas the latter got 0.88 F1-score.
This is understandable because, in the training phase, CAg only puts a sentence at 1 when all annotators agree, while MAg connotates 1 when the majority agrees. Therefore, there is much more noise in MAg than in CAg.

The second experiment performed a comparison between several ML algorithms in order to understand which one or ones would get better results in classifying short messages. 
For this purpose, it was defined that the CAg annotation reading model would be used, along with testing the two sentence embeddings possibilities (raw sentence embeddings and feature reduced sentence embeddings).

\begin{table*}[h!]
    \vspace*{-1mm}
    \begin{center}
    \def\arraystretch{1.5}%
        \begin{tabular}{l|l||c|c|c|c|c|c|c}
            \toprule
            \multicolumn{1}{c|}{\multirow{2}{*}{Data}} &\multicolumn{1}{c||}{\multirow{2}{*}{Metrics}} & \multicolumn{7}{c}{Machine Learning Algorithms} \\\cline{3-9}
            && MLP & BNB & KNN & XGB & GNB & SVM & LR \\\hline
            \multicolumn{1}{c|}{\parbox[t]{5mm}{\multirow{4}{*}{\rotatebox[origin=c]{90}{\makecell[tl]{Without \\Reduction}}}}} & Accuracy & 0.946 & 0.940 & 0.930 & 0.950 & 0.945 & \textbf{0.957} & 0.953  \\\cline{2-9}
            \multicolumn{1}{c|}{} & Precision  & 0.940 & 0.939 & 0.895 & 0.952 & 0.950 & \textbf{0.964} & 0.956 \\\cline{2-9}
            \multicolumn{1}{c|}{} & Recall  & 0.955 & 0.944 & \textbf{0.976} & 0.950 & 0.941 & 0.950 & 0.951  \\\cline{2-9}
            \multicolumn{1}{c|}{}& F1-score  & 0.946 & 0.940 & 0.932 & 0.949 & 0.944 & \textbf{0.956} & 0.952 
             \\\hline\hline
            \multicolumn{1}{c|}{\parbox[t]{5mm}{\multirow{4}{*}{\rotatebox[origin=c]{90}{\makecell[tl]{With \\Reduction}}}}} & Accuracy & 0.948 & 0.941 & 0.943 & 0.950 & 0.944 & \textbf{0.955} & 0.948  \\\cline{2-9}
            \multicolumn{1}{c|}{} & Precision  & 0.945 & 0.952 & 0.920 & 0.950 & \textbf{0.961} & 0.960 & 0.955  \\\cline{2-9}
            \multicolumn{1}{c|}{}& Recall  & 0.948 & 0.941 & 0.943 & 0.950 & 0.944 & \textbf{0.955} & 0.948 \\\cline{2-9}
            \multicolumn{1}{c|}{}& F1-score  & 0.949 & 0.941 & 0.945 & 0.950 & 0.944 & \textbf{0.955} & 0.948
            \\
            \bottomrule
        \end{tabular}
    \end{center}
    \caption{Accuracy, precision, recall and f1-score performances for different set of machine learning algorithms, with and without feature reduction.}
    \label{table:comparison_results_table}
\end{table*}

\begin{figure}[!h]
	\centering
	\subfloat[F1-score Normal Distribution without features reduced]{%
		\includegraphics[width=0.8\linewidth]{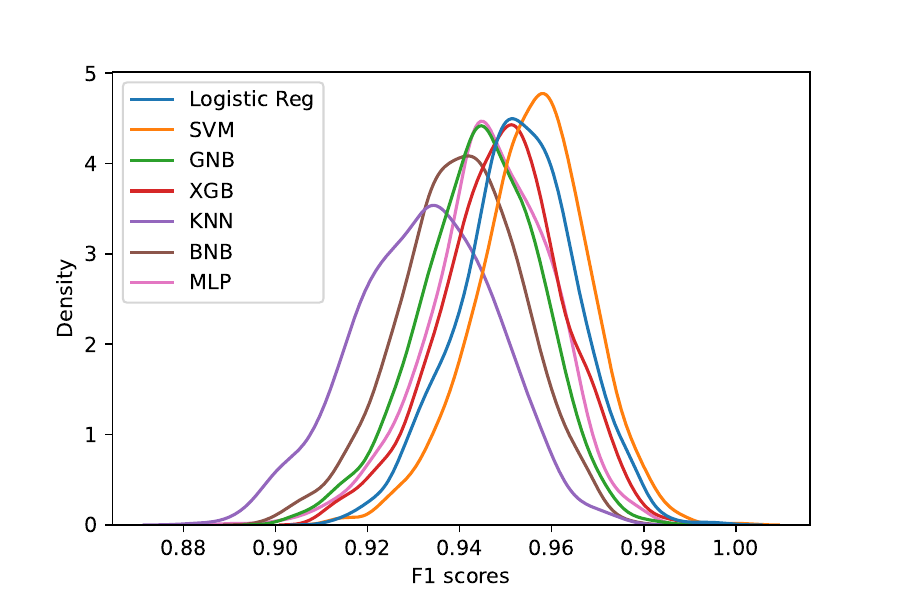}}
	\\
	\subfloat[F1-score Normal Distribution with features reduced]{%
		\includegraphics[width=0.8\linewidth]{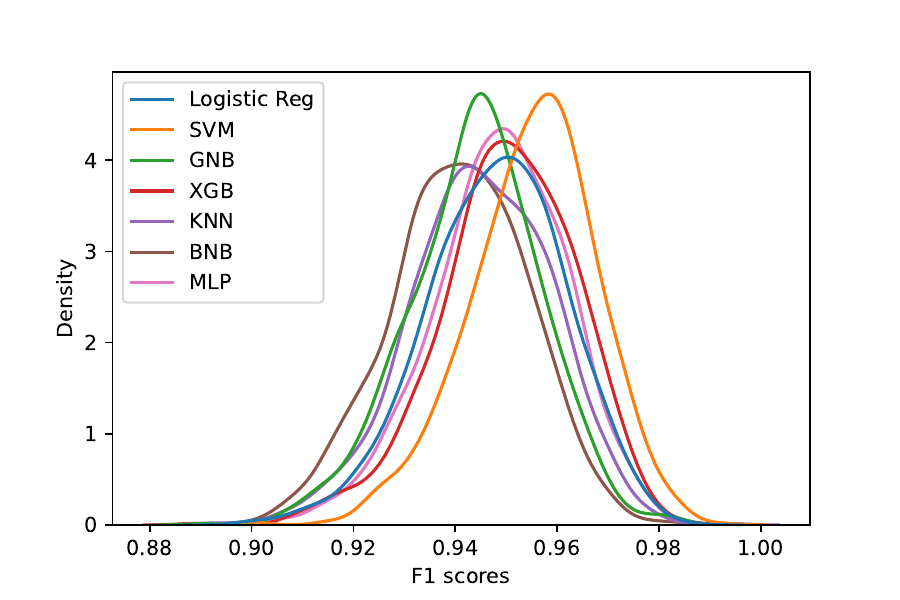}}
		\caption{Comparison between different Machine Learning Models: a) without features reduced and b) features reduced.}
	\label{fig:comp_models} 
\end{figure}

As we can see in both Figure~\ref{fig:comp_models} and Table~\ref{table:comparison_results_table}, after 1000 MC's simulations, the values of the two models are similar.
It is also visible that all tested algorithms obtained average results above 0.93 for all (accuracy, precision, recall, f1-score) evaluation metrics and that the most prominent algorithm, for both approaches was the Support Vector Machine.
%


%
The first approach had no embedding reduction, whereas the second one got embeddings randomly reduced. They both turned out with a median and average f1-scores values of 0.956 and 0.955, respectively.
\begin{figure}[!h]
	\centering
	\includegraphics[width=0.8\linewidth]{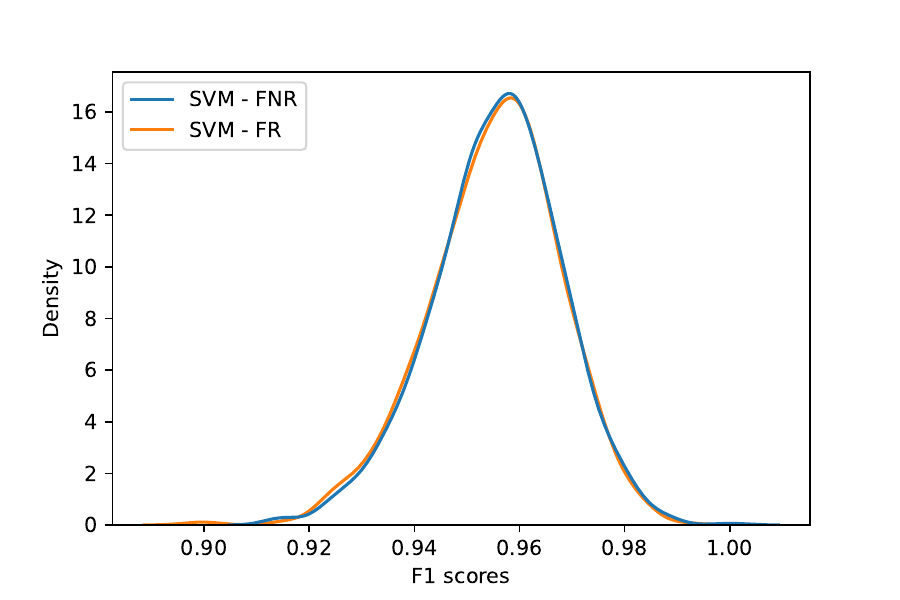}
	\caption{Comparison between SVM results with (FR) and without (FNR) features reduced.}
	\label{fig:comp_svm_approaches} 
\end{figure}
In Figure~\ref{fig:comp_svm_approaches}, it is possible to compare the behaviour of these two approaches (embeddings with and without reduction) for the SVM algorithm over the 1000 MC simulations.

\begin{figure}[!h]
    \centering
    \includegraphics[width=0.8\linewidth]{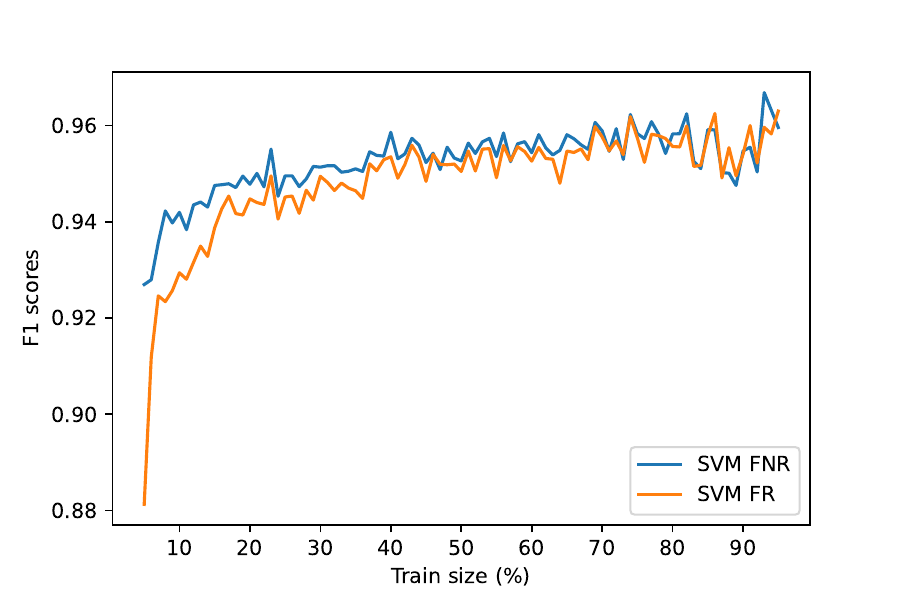}
    \caption{Train Size Analysis.}
	\label{fig:train_size}
\end{figure}

After reaching such positive results for the randomly selected features model, we studied how the amount of training data in a estimator would affect its results. 
For this matter, the CAg annotation reading model was selected, as well as SVM algorithm that had obtained the best results.
We determined a training dataset minimum and maximum of 5\% and 95\%, respectively, and obtained the results illustrated in figure~\ref{fig:train_size}.

We can observe that, even with a low amount of training data, we achieved optimal results regarding the classification of text messages. We also obtained an average f1-score results of 0.94 for training data between 15\% and 25\%.

Finally, in order to evaluate the model, we performed cross-validation~\cite{crossValidation}.
Cross-validation is used to evaluate the performance of the estimator and allows the model to learn and be tested on different data. 
This is important because a model that simply repeated the labels of the samples it had just seen, would produce a perfect result, but it would not make useful predictions for data not yet seen.
Thus, we defined, once again, that the annotation reading type would be CAg.
The ML algorithm would be SVM and that the sentence embeddings features would be randomly selected, as previously explained. 
The number of re-shuffling and splitting iterations was set to 10 and the number of training data to 20\% of the dataset.
With scikit-learn library's help to calculate the cross-validation, we obtained the average result of 0.95 f1-score with a standard deviation of 0.01.

\section{Conclusion and Future Work}

Our study demonstrates that it is feasible to use SBERT as a feature for classifying short messages in online chat conversations. 
This research aims to aid social science researchers and educators in gauging the level of engagement of online chat participants on a particular subject. 
Although our pipeline was developed using only one theme, we believe that it has potential for incorporation into future work, in other subjects. 
Our results suggest that utilizing BERT-based techniques to classify online chat room messages from online conversations can considerably enhance machine classification outcomes. 
Additionally, we have demonstrated that reducing the number of embeddings features by approximately
85\% can produce similar outcomes to training algorithms with raw embeddings with 768 dimensions. 
Lastly, we have proven that by training machine learning algorithms with a smaller percentage of training data (approximately 20\% of the dataset), we can achieve results that surpass our expectations: an average f1-score of 0.94 with a standard deviation of 0.01.

This work will be continued with further developments on the modules presented in this paper.
Other techniques, like Deep learning are very promising in terms of further supporting social scientists in better understanding human communication and persuasion in online chat rooms. 
Although its use may have significant limitations in a few categories, the overall advances are encouraging.
As guidelines for future work, the list below enumerates some of the main topics that will provide
novel contributions:
\begin{itemize}
    \item Chat rooms messages temporal analysis with deep learning temporal networks.
    \item Turn shift analysis, during the debate.
    \item An increase in the dataset size for other ages and data types sources.
\end{itemize}

\section*{Acknowledgment}
This research was partially funded by Funda\c{c}\~ao para a Ci\^{e}ncia e a Tecnologia under Projects "Factors for promoting dialogue and healthy behaviours in online school communities" with reference DSAIPA/DS/0102/2019 and developed at the R\&D Unit CICANT - Research Center for Applied Communication, Culture and New Technologies, UIDB/04111/2020, UIDB/50008/2020 as well as Instituto Lus\'ofono de Investiga\c{c}\~ao e Desenvolvimento (ILIND) under Project COFAC/ILIND/COPELABS/1/2022.

%
%
\bibliography{IEEEabrv,main}

\begin{thebibliography}{10}
\providecommand{\url}[1]{#1}
\csname url@samestyle\endcsname
\providecommand{\newblock}{\relax}
\providecommand{\bibinfo}[2]{#2}
\providecommand{\BIBentrySTDinterwordspacing}{\spaceskip=0pt\relax}
\providecommand{\BIBentryALTinterwordstretchfactor}{4}
\providecommand{\BIBentryALTinterwordspacing}{\spaceskip=\fontdimen2\font plus
\BIBentryALTinterwordstretchfactor\fontdimen3\font minus \fontdimen4\font\relax}
\providecommand{\BIBforeignlanguage}[2]{{%
\expandafter\ifx\csname l@#1\endcsname\relax
\typeout{** WARNING: IEEEtran.bst: No hyphenation pattern has been}%
\typeout{** loaded for the language `#1'. Using the pattern for}%
\typeout{** the default language instead.}%
\else
\language=\csname l@#1\endcsname
\fi
#2}}
\providecommand{\BIBdecl}{\relax}
\BIBdecl

\bibitem{Careaga}
\BIBentryALTinterwordspacing
M.~Careaga-Butter, B.~Q. María~Graciela, and F.-H. Carolina, ``Critical and prospective analysis of online education in pandemic and post-pandemic contexts: Digital tools and resources to support teaching in synchronous and asynchronous learning modalities,'' \emph{Aloma: revista de psicologia, ciències de l’educació i de l’esport Blanquerna}, vol.~38, no.~2, pp. 23--32, dic. 2020. [Online]. Available: \url{https://raco.cat/index.php/Aloma/article/view/377756}
\BIBentrySTDinterwordspacing

\bibitem{Uthus2013}
D.~C. Uthus and D.~W. Aha, ``Multiparticipant chat analysis: A survey,'' pp. 106--121, 2013.

\bibitem{Anjewierden2007}
A.~Anjewierden, B.~Kollöffel, and C.~Hulshof, ``Towards educational data mining: Using data mining methods for automated chat analysis to understand and support inquiry learning processes,'' 2007.

\bibitem{Trausan}
\BIBentryALTinterwordspacing
S.~Trausan-Matu, T.~Rebedea, J.~Fong, F.~L. Wang, A.~Dragan, and C.~Alexandru, ``Visualisation of learners' contributions in chat conversations,'' pp. 217--226, 2007. [Online]. Available: \url{https://www.researchgate.net/publication/210241895}
\BIBentrySTDinterwordspacing

\bibitem{Alsmadi2019}
I.~Alsmadi and K.~H. Gan, ``Review of short-text classification,'' pp. 155--182, 6 2019.

\bibitem{Danilov2021}
G.~Danilov, T.~Ishankulov, K.~Kotik, Y.~Orlov, M.~Shifrin, and A.~Potapov, ``The classification of short scientific texts using pretrained bert model,'' pp. 83--87, 7 2021.

\bibitem{Demirsoz2017}
O.~Demirsoz and R.~Ozcan, ``Classification of news-related tweets,'' \emph{Journal of Information Science}, vol.~43, pp. 509--524, 8 2017.

\bibitem{Hu2022}
Y.~Hu, J.~Ding, Z.~Dou, and H.~Chang, ``Short-text classification detector: A bert-based mental approach,'' \emph{Computational Intelligence and Neuroscience}, vol. 2022, 2022.

\bibitem{Lee2016}
\BIBentryALTinterwordspacing
J.~Y. Lee and F.~Dernoncourt, ``Sequential short-text classification with recurrent and convolutional neural networks,'' 3 2016. [Online]. Available: \url{http://arxiv.org/abs/1603.03827}
\BIBentrySTDinterwordspacing

\bibitem{Kowsari2019}
K.~Kowsari, K.~J. Meimandi, M.~Heidarysafa, S.~Mendu, L.~Barnes, and D.~Brown, ``Text classification algorithms: A survey,'' 2019.

\bibitem{Devlin2018}
\BIBentryALTinterwordspacing
J.~Devlin, M.-W. Chang, K.~Lee, K.~T. Google, and A.~I. Language, ``Bert: Pre-training of deep bidirectional transformers for language understanding,'' 2018. [Online]. Available: \url{https://github.com/tensorflow/tensor2tensor}
\BIBentrySTDinterwordspacing

\bibitem{Lin2020}
Y.~H. Lin, C.~Y. Chen, J.~Lee, Z.~Li, Y.~Zhang, M.~Xia, S.~Rijhwani, J.~He, Z.~Zhang, X.~Ma, A.~Anastasopoulos, P.~Littell, and G.~Neubig, ``Choosing transfer languages for cross-lingual learning,'' \emph{ACL 2019 - 57th Annual Meeting of the Association for Computational Linguistics, Proceedings of the Conference}, pp. 3125--3135, 2020.

\bibitem{sbert}
\BIBentryALTinterwordspacing
N.~Reimers and I.~Gurevych, ``Sentence-bert: Sentence embeddings using siamese bert-networks,'' \emph{CoRR}, vol. abs/1908.10084, 2019. [Online]. Available: \url{http://arxiv.org/abs/1908.10084}
\BIBentrySTDinterwordspacing

\bibitem{Hidey2017}
C.~Hidey, E.~Musi, A.~Hwang, S.~Muresan, and K.~McKeown, ``Analyzing the semantic types of claims and premises in an online persuasive forum,'' pp. 11--21, 2017.

\bibitem{Meredith2014}
J.~Meredith and E.~Stokoe, ``Repair: Comparing facebook 'chat' with spoken interaction,'' \emph{Discourse and Communication}, vol.~8, pp. 181--207, 2014.

\bibitem{xuan2019}
H.~X. Huynh, V.~T. Nguyen, N.~Duong-Trung, V.~H. Pham, and C.~T. Phan, ``Distributed framework for automating opinion discretization from text corpora on facebook,'' \emph{IEEE Access}, vol.~7, pp. 78\,675--78\,684, 2019.

\bibitem{Jucker2021}
A.~H. Jucker, ``Methodological issues in digital conversation analysis,'' 8 2021.

\bibitem{Meredith2019}
\BIBentryALTinterwordspacing
J.~Meredith, ``Conversation analysis and online interaction,'' \emph{Research on Language and Social Interaction}, vol.~52, pp. 241--256, 2019. [Online]. Available: \url{https://doi.org/10.1080/08351813.2019.1631040}
\BIBentrySTDinterwordspacing

\bibitem{Paulus2016}
\BIBentryALTinterwordspacing
T.~Paulus, A.~Warren, and J.~N. Lester, ``Applying conversation analysis methods to online talk: A literature review,'' \emph{Discourse, Context and Media}, vol.~12, pp. 1--10, 2016. [Online]. Available: \url{http://dx.doi.org/10.1016/j.dcm.2016.04.001}
\BIBentrySTDinterwordspacing

\bibitem{Liu2022}
Y.~Liu, P.~Li, and X.~Hu, ``Combining context-relevant features with multi-stage attention network for short text classification,'' \emph{Computer Speech and Language}, vol.~71, 1 2022.

\bibitem{Gupta}
S.~Gupta, S.~E. Bolden, J.~Kachhadia, A.~Korsunska, and J.~Stromer-Galley, ``Polibert: Classifying political social media messages with bert,'' 2020.

\bibitem{Khatri2020}
A.~Khatri and A.~Kumar, ``Sarcasm detection in tweets with bert and glove embeddings,'' 2020.

\bibitem{Ye2020}
Z.~Ye, G.~Jiang, Y.~Liu, Z.~Li, and J.~Yuan, ``Document and word representations generated by graph convolutional network and bert for short text classification,'' vol. 325.\hskip 1em plus 0.5em minus 0.4em\relax IOS Press BV, 8 2020, pp. 2275--2281.

\bibitem{landis}
\BIBentryALTinterwordspacing
J.~R. Landis and G.~G. Koch, ``The measurement of observer agreement for categorical data,'' \emph{Biometrics}, vol.~33, no.~1, pp. 159--174, 1977. [Online]. Available: \url{http://www.jstor.org/stable/2529310}
\BIBentrySTDinterwordspacing

\bibitem{Krippendorff2016}
\BIBentryALTinterwordspacing
K.~Krippendorff, Y.~Mathet, S.~Bouvry, and A.~Widlöcher, ``On the reliability of unitizing textual continua: Further developments,'' \emph{Quality \& Quantity}, vol.~50, pp. 2347--2364, 11 2016. [Online]. Available: \url{http://link.springer.com/10.1007/s11135-015-0266-1}
\BIBentrySTDinterwordspacing

\bibitem{mikolov}
Y.~Goldberg and O.~Levy, ``word2vec explained: deriving mikolov et al.'s negative-sampling word-embedding method,'' \emph{arXiv preprint arXiv:1402.3722}, 2014.

\bibitem{Bert}
\BIBentryALTinterwordspacing
J.~Devlin, M.~Chang, K.~Lee, and K.~Toutanova, ``{BERT:} pre-training of deep bidirectional transformers for language understanding,'' \emph{CoRR}, vol. abs/1810.04805, 2018. [Online]. Available: \url{http://arxiv.org/abs/1810.04805}
\BIBentrySTDinterwordspacing

\bibitem{levy}
O.~Levy and Y.~Goldberg, ``Dependency-based word embeddings,'' in \emph{Proceedings of the 52nd Annual Meeting of the Association for Computational Linguistics (Volume 2: Short Papers)}, 2014, pp. 302--308.

\bibitem{umap}
\BIBentryALTinterwordspacing
L.~McInnes, J.~Healy, and J.~Melville, ``Umap: Uniform manifold approximation and projection for dimension reduction,'' 2018. [Online]. Available: \url{https://arxiv.org/abs/1802.03426}
\BIBentrySTDinterwordspacing

\bibitem{stoppiglia2003ranking}
H.~Stoppiglia, G.~Dreyfus, R.~Dubois, and Y.~Oussar, ``Ranking a random feature for variable and feature selection,'' \emph{The Journal of Machine Learning Research}, vol.~3, pp. 1399--1414, 2003.

\bibitem{xgboost_Chen_2016}
\BIBentryALTinterwordspacing
T.~Chen and C.~Guestrin, ``{XGBoost},'' in \emph{Proceedings of the 22nd {ACM} {SIGKDD} International Conference on Knowledge Discovery and Data Mining}.\hskip 1em plus 0.5em minus 0.4em\relax {ACM}, aug 2016. [Online]. Available: \url{https://doi.org/10.1145\%2F2939672.2939785}
\BIBentrySTDinterwordspacing

\bibitem{9849888}
G.~Mestre, J.~P. Matos-Carvalho, and R.~M. Tavares, ``Irrigation management system using artificial intelligence algorithms,'' in \emph{2022 International Young Engineers Forum (YEF-ECE)}, 2022, pp. 69--74.

\bibitem{svm}
\BIBentryALTinterwordspacing
N.~Cristianini and E.~Ricci, \emph{Support Vector Machines}.\hskip 1em plus 0.5em minus 0.4em\relax Boston, MA: Springer US, 2008, pp. 928--932. [Online]. Available: \url{https://doi.org/10.1007/978-0-387-30162-4\_415}
\BIBentrySTDinterwordspacing

\bibitem{2019}
\BIBentryALTinterwordspacing
J.~P. Matos-Carvalho, F.~Moutinho, A.~B. Salvado, T.~Carrasqueira, R.~Campos-Rebelo, D.~Pedro, L.~M. Campos, J.~M. Fonseca, and A.~Mora, ``Static and dynamic algorithms for terrain classification in uav aerial imagery,'' \emph{Remote Sensing}, vol.~11, no.~21, p. 2501, Oct 2019. [Online]. Available: \url{http://dx.doi.org/10.3390/rs11212501}
\BIBentrySTDinterwordspacing

\bibitem{crossValidation}
\BIBentryALTinterwordspacing
M.~Stone, ``Cross-validatory choice and assessment of statistical predictions,'' \emph{Journal of the Royal Statistical Society. Series B (Methodological)}, vol.~36, no.~2, pp. 111--147, 1974. [Online]. Available: \url{http://www.jstor.org/stable/2984809}
\BIBentrySTDinterwordspacing

\end{thebibliography}
\bibliographystyle{IEEEtran}

\end{document}